\documentclass[conference]{IEEEtran}
\IEEEoverridecommandlockouts
\usepackage{todonotes}
\usepackage{cite}
\usepackage{amsmath,amssymb,amsfonts}
\usepackage{algorithmic}
\usepackage{algorithm}
\usepackage{graphicx}
\usepackage{textcomp}
\usepackage{xcolor}
\def\BibTeX{{\rm B\kern-.05em{\sc i\kern-.025em b}\kern-.08em
    T\kern-.1667em\lower.7ex\hbox{E}\kern-.125emX}}
\begin{document}

\title{Predictability-Based Curiosity-Guided Action Symbol Discovery\\

\thanks{This work was supported by the Scientific and Technological Research Council of Turkey (TUBITAK) ARDEB 1001 Program (124E227) and by the INVERSE
Project under Grant 101136067, funded by the European Union.} 
}

\author{\IEEEauthorblockN{Burcu Kilic}
\IEEEauthorblockA{\textit{Department of Computer Engineering} \\
\textit{Bogazici University}\\
Istanbul, Turkey \\
burcu.kilic@std.bogazici.edu.tr}
\and
\IEEEauthorblockN{Alper Ahmetoglu}
\IEEEauthorblockA{\textit{Intelligent Robot Lab} \\
\textit{Brown University}\\
Providence, Rhode Island, US\\
aahmetog@cs.brown.edu}
\and
\IEEEauthorblockN{Emre Ugur}
\IEEEauthorblockA{\textit{Department of Computer Engineering} \\
\textit{Bogazici University}\\
Istanbul, Turkey \\
emre.ugur@bogazici.edu.tr}

}

\maketitle

\begin{abstract}
Discovering symbolic representations for skills is essential for abstract reasoning and efficient planning in robotics. Previous neuro-symbolic robotic studies mostly focused on discovering perceptual symbolic categories given a pre-defined action repertoire and generating plans with given action symbols. A truly developmental robotic system, on the other hand, should be able to discover all the abstractions required for the planning system with minimal human intervention. In this study, we propose a novel system that is designed to discover symbolic action primitives along with perceptual symbols autonomously. Our system is based on an encoder-decoder structure that takes object and action information as input and predicts the generated effect. To efficiently explore the vast continuous action parameter space, we introduce a Curiosity-Based exploration module that selects the most informative actions---the ones that maximize the entropy in the predicted effect distribution. The discovered symbolic action primitives are then used to make plans using a symbolic tree search strategy in single- and double-object manipulation tasks. We compare our model with two baselines that use different exploration strategies in different experiments. The results show that our approach can learn a diverse set of symbolic action primitives, which are effective for generating plans in order to achieve given manipulation goals.
\end{abstract}


\section{Introduction}

Humans and animals have the ability to perform abstract reasoning about their environments by learning abstract representations of actions and objects. For this, infants acquire high-level skills primarily by exploring their environment out of intrinsic curiosity, without an extrinsic end goal or rewards. Inspired by human developmental processes, we aim to demonstrate such autonomous exploration and high-level skill discovery in robotic agents. We present a robotic agent that can learn object and action abstractions by predicting the effects of its actions and exploring while trying to reduce its uncertainty in the predictions; in other words, exploring out of curiosity.

Symbolic representations of skills can enhance a robot's ability to reason and plan \cite{ugur2025,konidaris2019necessity}. Abstracting continuous sensorimotor information into discrete symbolic representations can simplify complex decision-making processes. These abstractions enable a robot to have better generalizable and transferable skills. Symbolic action primitives are utilized in symbolic planning, an effective method for efficiently generating action sequences to reach a specific goal state. 

Previous neuro-symbolic robotic approaches \cite{ahmetoglu2022deepsym,ahmetoglu2022learning,ahmetoglu2024discovering} have shown success in learning object categories and relational object symbols from the bottleneck layer of an effect prediction encoder-decoder network. In \cite{ahmetoglu2025symbolic}, symbolic planning was performed using Planning Domain Description Language (PDDL)\cite{aeronautiques1998pddl} with learned object symbols and a set of predefined abstract actions. However, these studies depend on manually defined discrete actions, and there is no autonomous skill discovery part.

A real cognitive developmental system, on the other hand, should have the ability to discover high-level discrete actions as well. In an early study, a robot was initialized with a basic reach-and-grasp movement capability, discovering a set of action primitives by exploring its action parameter space and applying clustering in its tactile measurements  \cite{ugur2015staged,ugur2012self}. However, the learned primitives were not used in high-level planning. More recently, \cite{silver2021,chitnis2022} framed the problem as an operator learning problem in their Task and Motion Planning (TAMP) framework. They proposed a neuro-symbolic relational transition model where a task plan was generated using symbolic planning, and then a neural network was used to search for the low-level operator parameters during execution. \cite{silver2022a} learned action primitives, providing a bi-level operator learning stack. However, these methods followed an algorithmic approach in learning operators, whereas we proposed to use a generic neural network for object and action symbol discovery.

The studies reviewed above mostly relied on random exploration. It is known that infants benefit from intrinsic motivation to more efficiently explore their environment \cite{oudeyer2016intrinsic}. Similar ideas have been applied in artificial intelligence and robotics. \cite{houthooft2016vime} provided rewards to agents for actions that give high information gain, \cite{seo2021state} used the state entropy as an intrinsic reward, \cite{lopes2012exploration} used learning progress (LP) \cite{schmidhuber1991possibility} for exploration guiding in reinforcement learning, \cite{bugur2019effect, Sener2023} used LP for exploration region selection in object-action-effect spaces, and   \cite{pathak2017curiosity} used the error from the effect prediction model as a reward signal to the action policy network. In our study, we propose to select actions that maximize the entropy in our model, which predicts the effects of actions.

The aim of this work is to find discrete action primitives that are effective in planning. For this, significantly extending our object symbol discovery framework \cite{ahmetoglu2022deepsym}, we propose a predictive encoder-decoder neural network, which takes action parameters and object features as input and generates action effects as output, is proposed. The core idea is to binarize the embeddings in the bottleneck layer, enabling symbolic planning. We also propose to guide the exploration of the robot by a curiosity signal, which depends on the entropy of the neural network output activation. Finally, after a sequence of actions is generated via a tree-based symbolic planner, the continuous parameters of the corresponding action are found using a gradient-based method, which freezes network weights and applies a search in the action parameter space. 

We used a manipulation robot with a gripper in a simulated environment for experimental verification. Exploring the parameter space of a reach action, our method discovered a diverse set of action primitives such as grasp, grasp and place, pull, and push in different directions. We used these primitives to generate symbolic plans to bring an object to a goal position and execute the actions with the proposed parameter distillation procedure. We also showed that our curiosity-based exploration overperformed the baselines, using other exploration strategies.
In the rest of this paper, we first provide the details of our proposed method, then give the experimental setup and the baselines, and finally provide the experimental results. 

\section{Proposed Method}

\begin{figure}
    \centering
    \includegraphics[width=0.85\linewidth]{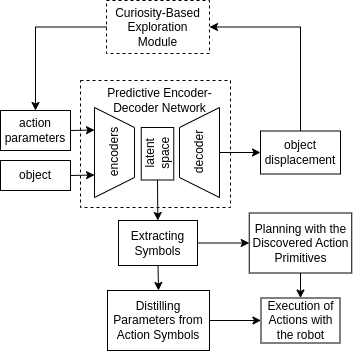}
    \caption{Overview of the proposed method. The effect prediction model is an encoder-decoder deep neural network that predicts a distribution over the effect. The entropy of the distribution is given to the curiosity-based exploration module to guide the action selection process. The object is randomly initialized in the environment. The symbols are generated via binarization of the bottleneck layer of the effect prediction model.}
    \label{fig:method}
\end{figure}

We provide the sensorimotor representation, followed by the predictive encoder-decoder network, continuous action parameterization, curiosity-based exploration module, and the planner. An overview of our proposed method can be seen in Figure \ref{fig:method}.


\subsection{Sensorimotor Representation}
Each object $o \in \mathbb{R}^4$ is represented as $o=[s_x, s_y, d, t]$ where $s_x, s_y, d$ denotes the object's dimensions in different axes, and $t$ corresponds to the type of the object. A state $s \in \mathbb{R}^{4 \times m}$ is the state of the environment with $m$ objects. The object features, which are used as the input of the predictive neural network, correspond to the target object's features.

The robotic action $a \in \mathbb{R}^{12}$ is a continuous vector concatenating the start, middle, and end points of the robot's trajectory.
\begin{equation}\label{action_params}
    a = [p_1, p_2, p_3], p_i = [x_i, y_i, z_i, g_i] \in \mathbb{R}^4, i=1,2,3.
\end{equation}

Here, $x_i, y_i, z_i$ denote the coordinates of the robot relative to the target object, and $g_i$ is a parameter that defines the state of the gripper. The gripper is open if $g_i>=0.5$ and closed otherwise. 

The effect $e \in \mathbb{R}^3$ of an action denotes the change in the target object's absolute position, $e = [\Delta x^o_t, \Delta y^o_t, \Delta z^o_t]$. 

\subsection{Discovering Symbols in Predictive Encoder-Decoder Network}\label{subsec:effect_prediction_model}

\subsubsection{Extracting Symbols} In our effect prediction model, we aim to extract action and object symbols from the bottleneck layer of the encoder-decoder deep neural network. Specifically, we propose to learn a mapping $\phi_o:\mathbb{R}^4 \rightarrow{} \mathbb{R}^j$ in the object encoder and a mapping $\phi_a:\mathbb{R}^{12} \rightarrow{} \mathbb{R}^k$ in the action encoder, discretizing the continuous outputs of these encoders into discrete symbols using a binary step function. After binarization, we obtain discrete representations for both the action and the object:
\begin{equation}\label{discrete_symbols}
    z_o = b(\phi_o(o)) \in \{0,1\}^j, z_a = b(\phi_a(a)) \in \{0,1\}^k.
\end{equation}

\subsubsection{Predictive Encoder-Decoder Network Architecture} 
\begin{figure*}
    \centering
    \includegraphics[width=\textwidth]{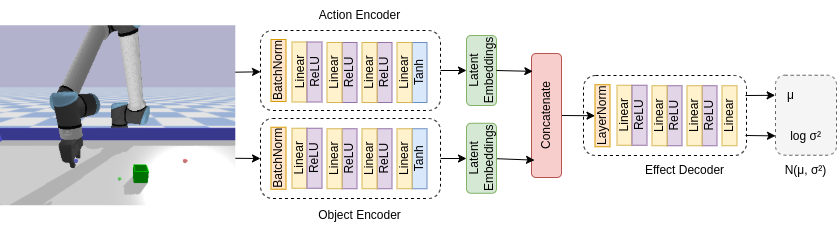}
    \caption{Overview of the proposed effect prediction model. The first image shows the PyBullet environment with an example object. The blue, green, and red dots show the start, middle, and end points of the robot's trajectory, respectively. The action and object encoders generate latent embeddings, which are then concatenated. The decoder uses these combined embeddings to predict a Gaussian distribution over the object displacement caused by the action. The symbols are then extracted from the latent embeddings with the given method in Section \ref{parameter_distill}.}
    \label{fig:deepsym}
\end{figure*}

The core of the architecture is a deep neural network with separate object and action encoders, followed by a single decoder, as shown in Figure \ref{fig:deepsym}. The network predicts a Gaussian distribution over the effect given action parameters and object features:

\begin{equation}\label{effect_distribution}
    p(e \mid a,o)=\mathcal{N}(e; \mu(a,o),\sigma^2(a,o))
\end{equation}

The action and object encoders have an initial Batch Normalization layer. They consist of four hidden layers, the first three of which are linear layers with ReLU activation, and the final layer, which is a linear layer with Tanh activation function. The \(\tanh\) activation function ensures the embeddings are between -1 and 1.

The decoder is a 4-layer perceptron with linear layers and ReLU activation in the hidden layers. It takes the concatenated object and action embeddings $z = [z_o, z_a]$. Layer Normalization is applied to the concatenated embeddings to ensure a stable distribution before 
further processing. In the final linear layer, the decoder predicts a mean $\mu$ and log variance $log(\sigma^2)$ for each of the x, y, and z axes, thereby creating three independent normal distributions over the predicted effect shown as Equation \ref{normal_distribution}. Dropout is applied to all the hidden layers in the model for regularization. The predicted effect as network output is calculated as follows.

\begin{equation}\label{normal_distribution}
    p(x \mid \mu, \sigma^2) = \frac{1}{\sqrt{2\pi\sigma^2}} \exp\left(-\frac{(x-\mu)^2}{2\sigma^2}\right).
\end{equation}

Based on the predicted and observed effects, the negative log-likelihood (NLL) loss is used:
\begin{equation}\label{nll_loss}
\mathcal{L}_{\text{NLL}}(x) = -\log p(x \mid \mu, \sigma^2)
= \frac{1}{2}\log(2\pi\sigma^2) + \frac{(x-\mu)^2}{2\sigma^2}.
\end{equation}

To effectively extract distinct symbols from the latent space, we need to push the embeddings of different actions from each other and pull similar ones together. For this, we used Normalized Temperature-Scaled Cross Entropy (NT-Xent) Loss \cite{chen2020simple}. The concatenated action and object embeddings are normalized for a batch of $N$ samples. With the normalized embeddings $\tilde{z}_i = \frac{z_i}{\|z_i\|}$ and temperature $\tau$, we define the similarity as follows:

\begin{equation}\label{similarity}
    s_{ij} = \tilde{z}_i^\top \tilde{z}_j.
\end{equation}

The NT-Xent loss is then

\begin{equation}\label{contrastive_loss}
\mathcal{L}_{\text{NT-Xent}} = -\frac{1}{N} \sum_{i=1}^{N} \log \frac{\exp\left(\frac{s_{ii}}{\tau}\right)}{\sum_{j=1}^{N}\exp\left(\frac{s_{ij}}{\tau}\right)},
\end{equation}

Finally, with a loss coefficient $\lambda$, the total loss is defined as
\begin{equation}\label{total_loss}
\mathcal{L}_{\text{total}} = \lambda \Big( \mathcal{L}_{\text{NLL}} + \mathcal{L}_{\text{NT-Xent}} \Big).
\end{equation}

After training the effect prediction model, $z_o$ and $z_a$ are found by binarizing the latent variables from the outputs of the encoders. 

\subsection{Planning with the Discovered Action Primitives} \label{subsec:bfs}

Breadth-first tree search (BFS) is used for planning where the branch factor corresponds to the number of discovered action primitives. The goal is to reach a given goal state $s_g$ (within an error threshold) from an initial state $s_i$ by an action sequence $\pi$. The states are defined by the absolute positions of all the objects in the environment. The planner iteratively expands the candidate sequences. If the state achieved by an action sequence $\pi$ is within an error margin of $s_g$, the sequence is accepted as a solution. If no suitable sequence is identified within a maximum search depth, the algorithm terminates, indicating a failure to generate the plan with the learned action primitives. After finding an action sequence, each action should be executed, and for this, their continuous parameters should be found. In the next section, we will provide the details of the parameter distillation procedure.


\subsection{Distilling Parameters from Discovered Action Symbols}\label{parameter_distill}

Inspired by \cite{aktas2024multi}, we use an optimization-based distillation process to convert the symbols to continuous action parameters. Initially, we start with the set of action parameters that were used to train the effect prediction model. These parameters are the initial estimates for the distilled action parameters. Using the model's trained action encoder, we generate the action embeddings for these action parameters. We freeze the action encoder's weights and optimize only the action parameters using Stochastic Gradient Descent (SGD). The optimization is to minimize the mean squared error (MSE) between the model's encoded action embedding and the target binary symbol. After convergence, we select the candidate action parameter with the lowest error as the distilled action primitive of action symbol $z_a$. This process inverts the mapping $\phi_a(a)$ and maps the discrete symbols to continuous parameters, which allows the planner to use these primitives during task execution for achieving the given task.

\subsection{Curiosity-Based Exploration Module}\label{subsec:curiosity}

The robot interacts in a continuous action space, where efficient exploration is critical. In order to efficiently explore the continuous action space and learn meaningful action primitives, we propose to use a curiosity-based exploration approach based on exploration by selecting actions that maximize entropy in the effect prediction. Our effect prediction model outputs a Gaussian distribution over the effect, and the entropy of this distribution represents the uncertainty of the model regarding the predicted effect of an action. Therefore, by maximizing the entropy of the effect distribution, the method prioritizes actions that the model is uncertain. This way, we aim to accelerate the learning process, encourage the discovery of diverse and effective action primitives, and show how an intrinsic reward signal can make the model learn instead of relying only on extrinsic rewards.

Our proposed algorithm for curiosity-based exploration is given in Algorithm \ref{alg:curiosity_algorithm}. In each exploration step, we uniformly sample a set of candidate action parameters. For each candidate, we forward it to our effect prediction model and obtain a distribution over the predicted effects. The entropy of the distributions for each dimension is calculated, the mean entropy of all dimensions is found, and the candidate action with the maximum mean entropy is selected. We follow a greedy policy, selecting this action $a$ to execute and save its effects to train the effect prediction model further. 
\begin{algorithm}[!htbp]
    \caption{Curiosity-Based Exploration}\label{alg:curiosity_algorithm}
    \begin{algorithmic}[1]
        \REQUIRE Object parameters \( o \in \mathbb{R}^4 \)
        \ENSURE Selected action \( a^* \in \mathbb{R}^{12} \)
        \STATE Set number of candidate actions: \( N\)
        \FOR{\( i = 1 \) \TO \( N \)}

            \STATE Uniformly sample action parameters\\ 
            $a^{(i)} \sim \mathcal{U}\bigl([-0.05, 0.05]^{12}\bigr)$

            \STATE Create normal distributions for the effect by forwarding the model\\
            $p(e \mid a^{(i)}, o) = \mathcal{N}\Bigl(e; \mu(a^{(i)}, o), \sigma^2(a^{(i)}, o)\Bigr)$
            
            \FOR{\( j = 1, 2, 3 \)}
                \STATE Calculate entropy for each axes x, y, z\\ 
                $H_j(a^{(i)}, o) = \frac{1}{2} \log\Bigl(2\pi e\, \sigma_j^2(a^{(i)}, o)\Bigr)$
                
            \ENDFOR
            \STATE Get the mean entropy\\ 
            $\bar{H}(a^{(i)}, o) = \frac{1}{3}\sum_{j=1}^{3} H_j(a^{(i)}, o)$
            
        \ENDFOR
        \STATE Select the action with the maximum entropy\\ 
        $a^* = \arg\max_{i \in \{1,\dots,N\}} \bar{H}(a^{(i)}, o)$
        
        \RETURN \( a^* \)
    \end{algorithmic}
\end{algorithm}

\section{Experiments}
The experiments are performed using a UR10 manipulation robot in a PyBullet simulation environment. In this section, we detail the experimental setup, evaluation, and comparisons with planning.

\subsection{Experimental Setup}
The encoders of our network have four hidden layers with 128 nodes per layer. Tanh activation function is used to constrain the embeddings within the range [-1, 1]. The object encoder's output dimension was set to 2-bits, allowing up to $2^2 = 4$ distinct object categories, while the action encoder produces 3-bit output, allowing at most $2^3=8$ distinct action primitives. The concatenated embeddings are normalized using Layer Normalization and processed through the decoder, which has 4 hidden layers consisting of 128 nodes per layer. The model is trained with a mini-batch size of 512, a learning rate of 1e-5, and gradient clipping. The loss is the sum of $\mathcal{L}_{\text{NLL}}$ and $\mathcal{L}_{\text{NT-Xent}}$ with a loss coefficient $\lambda = 0.01$.

During each step of our Curiosity-Based Exploration, an object with a random size and type (hollow or non-hollow) is initialized. Using Algorithm \ref{alg:curiosity_algorithm}, from a set of 2000 randomly sampled candidate action parameters, the action sequence $a^*$ that maximizes the effect prediction entropy is selected and executed. The displacement of the object is saved as the effect. At every 512 steps, the effect prediction model is trained with the gathered $(a, o, e)$ tuples for 10 update epochs. The entire exploration process takes 10,000 steps. After exploration, distilled parameters for all action symbols are obtained with the algorithm in Section \ref{parameter_distill}. The parameters are executed on the robot to annotate the learned high-level action primitives. Then, using the learned high-level action primitives, we perform single- and multi-object plannings in Section \ref{subsec:planning}.

\subsection{Baselines}

To measure the performance of our Curiosity-Based Model, we compare our findings with two baseline models: Random Exploration Model and Active Learning Model\cite{kilic2025learning}. The Random Exploration Model is based on executing randomly sampled action parameters in randomly generated environments. The Active Learning Model is based on training the effect prediction model with only the actions that have high effects on the objects in the environment. It is a task-specific approach to avoid noise in the training dataset. Both approaches predict the effect directly instead of distribution, hence overlooking the uncertainties in the predicted effects.

\subsection{Model Prediction Error}
To evaluate our curiosity-based exploration module, a test dataset of 2400 samples generated with random exploration is used. Each sample includes random action parameters, a randomly generated environment state, and the effect of the action on the target object after execution. Interactions where the total effect is smaller than 0.8 along the three axes were excluded to reduce noise. In our curiosity-driven approach, the model predicts a distribution over the effect, and in this section, we use the mean of the distribution as the predicted effect. In the Random Exploration module and the Active Learning model, the output of the effect prediction model is used as the predicted effect, as they do not produce a distribution. We find the absolute error between this predicted effect and the ground truth effect separately for each dimension. When the model prediction errors are compared, as shown in Table \ref{tab:prediction_errors}, our curiosity-based exploration approach has lower errors on all dimensions, meaning that it can provide a better generalization to the unseen data compared to the baseline approaches.

\begin{table}[t]
    \caption{Model Prediction Errors in x, y, z Axes. Units are in meters.}
    \label{tab:prediction_errors}
    \centering
    \renewcommand{\arraystretch}{1.4}
    \resizebox{\columnwidth}{!}{%
        \begin{tabular}{c|c|c|c}
        & Curiosity-Based M. & Active Learning M. & Random Exploration M.\\
        \hline

        x & \textbf{0.0843} & 0.0917  & 0.1236  \\
        y & \textbf{0.0828} & 0.1090  & 0.1406  \\ 
        z & \textbf{0.1540} & 0.1556  & 0.2210  \\
          
        \end{tabular}
    }
    
\end{table}

\subsection{Discovered Action Primitives}
\begin{figure}
    \centering
    \includegraphics[width=\linewidth]{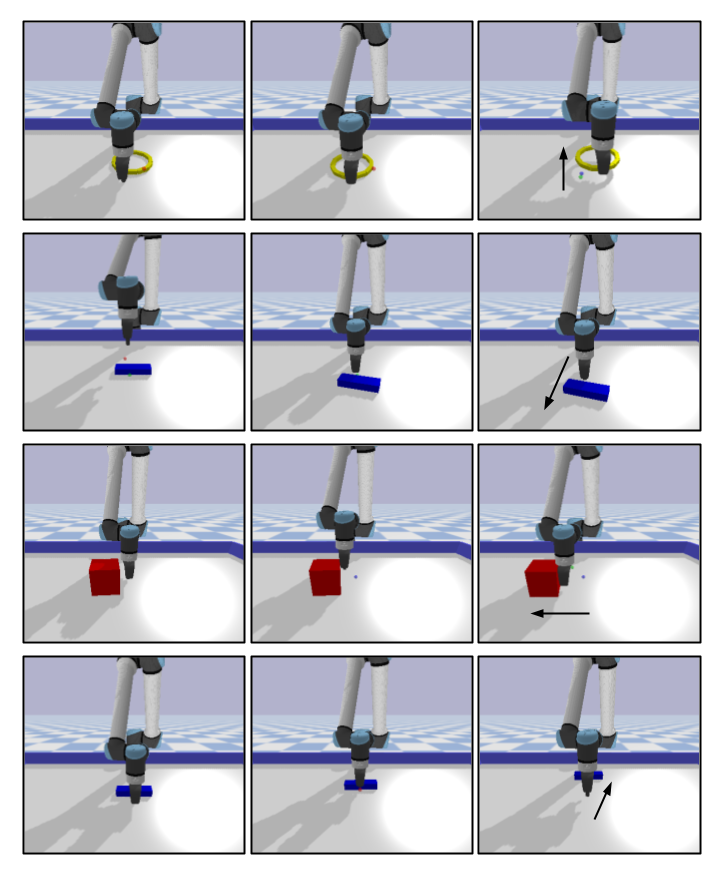}
    \caption{Example of the learned action primitives in the Curiosity-driven model. From top to bottom; pick up, forward push, left push, pull.}
    \label{fig:action_prims}
\end{figure}

After the exploration process, to understand and evaluate the diversity of the discovered action primitives, we first visually observe their prototypical executions and report their qualitative performance. To observe the executions of the action primitives, we distill the action parameters with the algorithm given in Section \ref{parameter_distill}. Some examples of high-level actions learned with our Curiosity-Based model are given in Figure \ref{fig:action_prims}.  To compare the learned primitives with other baselines, we performed the same experiment with the Active Learning Module and the Random Exploration Module that were explained earlier. The comparison is given in Table \ref{tab:learned_actions}. As shown, our method learned six different action primitives, including different push primitives, grasp, and release actions, whereas the baselines could discover four or three of these primitives. The random exploration module learned less meaningful actions since the continuous action space consists of mostly noisy and null actions. The curiosity-based model performed the best since it allows for selecting novel actions, resulting in a diverse set of high-level skills. With this, we can conclude that our Curiosity-Based Model has learned a richer set of high-level action primitives than the other two baseline models. 

\begin{table}[h]
    \caption{Learned Action Primitives}
    \label{tab:learned_actions}
    \centering
    \renewcommand{\arraystretch}{1.2}
    \resizebox{\columnwidth}{!}{%
        \begin{tabular}{c|c|c|c}
        & Curiosity-Based M. & Active Learning M. & Random Exploration M.\\
        \hline
        right push     & + & +  & +   \\
        left push      & + & +  & +   \\
        forward push   & + & +  & -   \\ 
        pull           & + & +  & -   \\
        grasp          & + & -  & +   \\
        pick and place & + & -  & -   \\
        \end{tabular}
    }
    
\end{table}

\subsection{Planning Performance}\label{subsec:planning}
In this section, we analyze the planning and plan execution performance of our model and provide a comparison with the baselines. For this, we generated random goal states and ran the planner to generate plans using learned action primitives. We consider single- and double-object planning tasks to verify the effectiveness of the learned action primitives. We generated 100 planning tasks, and for each task, a randomly initialized state, random action parameters, and the resulting state were recorded. In single-object tasks, the environment consists of only one object, and in double-object tasks, the state has two objects, and the planner should decide on the target object to execute an action primitive. Then, with the collected initial and goal state pairs, we perform the planning (with a maximum search depth of 3 and an error threshold of 0.05) in all three models. The success rate is defined as the percentage of plans that successfully reach the goal state. The success rates of our Curiosity-Based model, the Active Learning model, and the Random Exploration model are reported in Table \ref{tab:plan_ratio}. As shown, our model outperformed the baseline models, telling us that the learned high-level action primitives are more versatile. 

\begin{table}[]
    \caption{Successful Plan Ratio of the Models}
    \label{tab:plan_ratio}
    \centering
    \renewcommand{\arraystretch}{1.5}
    \resizebox{\columnwidth}{!}{%
        \begin{tabular}{c|c|c|c}
        & Curiosity-Based M. & Active Learning M. & Random Exploration M.\\
        \hline
        Single-Object & \textbf{82\%} & 56\%  & 13\%  \\
        Double-Object & \textbf{59\%} & 38\%  & 9\%  \\
        \end{tabular}
    }
\end{table}

\begin{figure}[ht!]
    \centering
    \includegraphics[width=\linewidth]{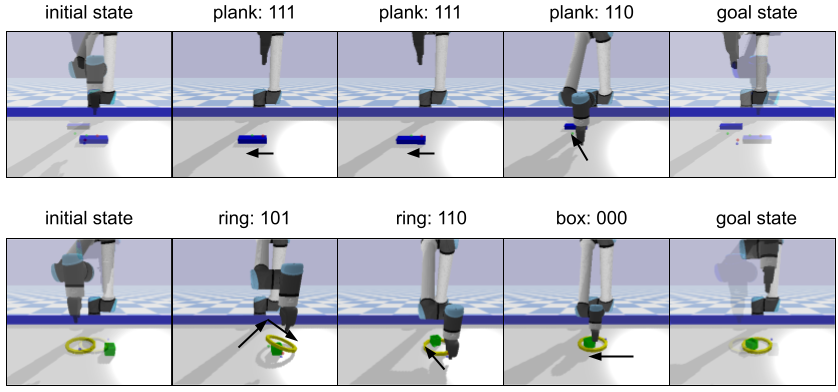}
    \caption{Example plans generated by BFS with the learned primitives in Curiosity-Based Model. The goal state is shown transparently in the first column, and the initial state is shown so in the last column to emphasize the effect.}
    \label{fig:example_plans}
\end{figure}

Figure \ref{fig:example_plans} shows sample plans generated by our planner with the learned primitives in the Curiosity-Based model. The first row shows a single-object plan consisting of left push and pull primitives, while the second row shows a two-object plan that has pick \& place, pull, and left push primitives.

\section{Conclusions}
In this study, we introduced a framework for discovering diverse high-level action primitives effective in next-state prediction and planning.  We proposed a deep neural network to predict a Gaussian distribution over the effect of an action on an object. To efficiently train the model, we designed a curiosity-based exploration module that selects the most informative actions (that maximize the entropy in the predicted effect distribution). We utilized the learned action and object symbols to perform single- and double-object manipulation tasks with a Breadth-First Search (BFS)-based planner. We showed that when the robot explores its environment using an entropy-based curiosity signal, compared to the random exploration module and effect-maximization, our method has better generalizing capability in predicting effects and learns a more useful and diverse set of meaningful high-level action primitives. We also showed that the action primitives found by our method can be effectively used by the symbolic planner in generating plans to achieve various single and paired object tasks and in executing these plans thanks to our parameter distillation approach.

In our work, we focused on the effects of a single object's displacement during exploration and training. Future work can extend this to include the relations between objects\cite{ahmetoglu2024discovering} to reason more comprehensively in multi-object tasks. Additionally, although the BFS-based planning strategy performed reasonably well in our experiments, it may work slower when new objects or more complex tasks are introduced. In the future, we plan to translate the learned abstractions and rules into Planning Domain Definition Language (PDDL)\cite{aeronautiques1998pddl} and use efficient off-the-shelf AI planners \cite{hoffmann2001ff,helmert2006fast}.


\bibliographystyle{IEEEtran}
\bibliography{ref}

\end{document}